\documentclass[10pt,twocolumn,letterpaper]{article}

\usepackage{cvpr-compact}
\usepackage{times}
\usepackage{epsfig}
\usepackage{graphicx}
\usepackage{lipsum}
\usepackage{amsmath}
\usepackage{amssymb}
\usepackage[numbers,sort]{natbib} 

\def\spacepreeqn{\vspace{-12pt}}
\def\spacetablecapt{\vspace{2pt}}
\def\spacesubsection{\vspace{-5pt}}
\def\spacesection{\vspace{-5pt}}

\usepackage[pagebackref=true,breaklinks=true,letterpaper=true,colorlinks,bookmarks=false]{hyperref}

\cvprfinalcopy 

\def\cvprPaperID{2873} 

\begin{document}

\title{Lip Reading Sentences in the Wild}

\author{Joon Son Chung$^1$\\
{\tt\scriptsize joon@robots.ox.ac.uk}
\and
Andrew Senior$^2$\\
{\tt\scriptsize andrewsenior@google.com}
\and
Oriol Vinyals$^2$\\
{\tt\scriptsize vinyals@google.com}
\and
Andrew Zisserman$^{1,2}$\\
{\tt\scriptsize az@robots.ox.ac.uk}
\and
$^1$Department of Engineering Science, University of Oxford 
\quad\quad $^2$Google DeepMind
}

\maketitle

\begin{abstract}
   The goal of this work is to recognise phrases
   and sentences being spoken by a talking face, 
   with or without the audio. 
   Unlike previous works that have
   focussed on recognising a limited number of words
   or phrases, we tackle lip reading as an 
   {\em open-world} problem -- unconstrained natural language sentences, and in the wild videos.

   Our key contributions are: 
   (1) a `Watch, Listen, Attend and Spell' (WLAS)  
   network that learns to transcribe videos of mouth motion
    to characters;
   (2) a curriculum learning strategy to accelerate training
   and to reduce overfitting; 
   (3) a `Lip Reading Sentences' (LRS)
    dataset for visual speech recognition, consisting of
   over 100,000 natural sentences from British television.

The WLAS model trained on the LRS dataset surpasses the performance of
all previous work on standard lip reading benchmark datasets, often by a
significant margin.  This 
lip reading performance beats a professional lip reader
on videos from BBC television, and we also demonstrate that visual
information helps to improve speech recognition performance even when
the audio is available.

\end{abstract}

\vspace{-10pt}
\section{Introduction}
\spacesection

Lip reading, the ability to recognize what is being said from visual
information alone, is an impressive skill, and very challenging
for a novice. It is inherently ambiguous at the word level due to 
homophemes --
different characters that produce exactly the same lip sequence ({\it e.g.}\
`p' and `b'). However, such ambiguities can be resolved to an
extent using the context of neighboring words in a sentence, and/or a
language model.

A machine that can lip read opens up a host of applications:
`dictating' instructions or messages to a phone in a noisy
environment; 
transcribing and re-dubbing archival silent films;  
resolving multi-talker simultaneous speech;
and, improving the 
performance of automated speech recogition in general.

That such automation is now possible is due to two developments that
are well known across computer vision tasks: the use of deep neural
network models~\cite{Krizhevsky12,Simonyan15,Szegedy15};
 and, the availability of a large scale dataset for
training~\cite{Russakovsky15,Larlus06}.  
In this case the model is based on the recent
sequence-to-sequence (encoder-decoder with attention) translater
architectures that have been developed for speech recognition and
machine
translation~\cite{graves2006connectionist,graves2014towards,bahdanau2014neural,chan2015listen,Sutskever14}. The
dataset developed in this paper is based on thousands of hours of
BBC television broadcasts that have talking faces together with
subtitles of what is being said.

We also investigate how lip reading can contribute to {\em audio}
based speech recognition. There is a large literature on this
contribution, particularly in noisy environments, as well as the
converse where some derived measure of audio can contribute to lip
reading for the deaf or hard of hearing. To investigate this aspect
we train a model to recognize characters from both audio and visual
input, and then systematically disturb the audio channel or remove the
visual channel.

Our model (Section~\ref{sec:arc}) outputs at the character level, is able to
learn a language model, and
has a novel dual attention mechanism that can operate over visual
input only, audio input only, or both. We show (Section~\ref{sec:training}) 
that training
can be accelerated by a form of curriculum learning.  We also describe
(Section~\ref{sec:dataset}) the generation and statistics of a new
large scale Lip Reading Sentences (LRS) 
dataset, based on BBC broadcasts containing talking faces
together with subtitles of what is said.  The broadcasts contain faces
`in the wild' with a significant variety of pose, expressions,
lighting, backgrounds, and ethnic origin.  This dataset will be
released as a resource for training and evaluation.

The performance of the model is assessed on a test set of the LRS
dataset, as well as on public benchmarks datasets for lip reading
including LRW~\cite{Chung16} and GRID~\cite{cooke2006audio}.  We
demonstrate {\em open world} (unconstrained sentences) lip reading on
the LRS dataset, and in all cases on public benchmarks the performance
exceeds that of prior work.

\subsection{Related works}
\spacesubsection

\noindent{\bf Lip reading.}
There is a large body of work on lip reading using 
pre-deep learning methods. 
These methods are thoroughly reviewed in
\cite{zhou2014review}, and we will not repeat this here.
A number of papers have used Convolutional Neural 
Networks (CNNs) to predict phonemes~\cite{noda2014lipreading}
 or visemes~\cite{koller2015deep}
from still images,
as opposed recognising to full words or sentences. 
A {\it phoneme} is the smallest distinguishable unit
of sound that collectively make up a spoken word;
 a {\it viseme} is its visual equivalent. 

For recognising full words,
Petridis {\it et al.}~\cite{petridisdeep} 
trains an LSTM classifier on 
a discrete cosine transform (DCT)
and deep bottleneck features (DBF).
Similarly, Wand {\it et al.}~\cite{wand2016lipreading}
uses an LSTM with HOG input features 
to recognise short phrases. 
The shortage of training data in lip reading 
presumably contributes to the continued use of
shallow features. 
Existing datasets consist of 
videos with only a small number of subjects,
and also a very limited vocabulary ($<$60 words),
which is also an obstacle to progress.
The recent paper of Chung and Zisserman~\cite{Chung16}
tackles the small-lexicon problem by using faces in
television broadcasts to assemble a dataset for 500 words.
However, as with any word-level 
classification task,
the setting is still distant from the
real-world, given that the word
boundaries must be known beforehand.
A very recent work~\cite{yannis2016lipnet} 
(under submission to ICLR 2017) uses a
CNN and LSTM-based network and Connectionist Temporal Classification 
(CTC)~\cite{graves2006connectionist} to compute the labelling.
This reports strong speaker-independent performance on the constrained
grammar and 51 word vocabulary of the GRID
dataset~\cite{cooke2006audio}. However, the method, suitably modified, should be
applicable to longer, more general sentences.

\noindent{\bf Audio-visual speech recognition.}
The problems of audio-visual speech recognition (AVSR)
and lip reading are
closely linked. 
Mroueh {\it et al.}~\cite{mroueh2015deep} employs
feed-forward Deep Neural Networks (DNNs) to perform
phoneme classification using a large
non-public audio-visual dataset.
The use of HMMs together with hand-crafted or pre-trained
visual features have proved popular
 --
 \cite{tamura2015audio} encodes
 input images using DBF; \cite{galatas2012audio} used DCT;
 and \cite{noda2015audio} uses a CNN pre-trained 
 to classify phonemes;
 all three combine these features with
 HMMs to classify spoken digits or isolated words.
 As with lip reading, there has been little
 attempt to develop AVSR systems that generalise
 to real-world settings.

\noindent{\bf Speech recognition.}
There is a wealth of literature on 
speech recognition systems 
 that utilise separate components for acoustic and
 language-modelling functions
({\em e.g.} hybrid DNN-HMM systems), 
that we will not review here.
  We restrict this
review to methods that
can be trained end-to-end.

For the most part, prior work can be divided into two types.
The first type uses CTC~\cite{graves2006connectionist}, where the model typically
predicts framewise labels and then looks for the optimal alignment
between the framewise predictions and the output sequence. 
The weakness is that the output labels are not conditioned on
each other.

The second type is sequence-to-sequence models~\cite{Sutskever14} that first read
all of the input sequence before starting to predict the output sentence.
A number of papers have adopted this approach
for speech recognition~\cite{chorowski2015attention,chorowski2014end}, and
the most related work to ours is that of Chan {\it et
al.}~\cite{chan2015listen} which proposes
an elegant sequence-to-sequence method
to transcribe audio signal to characters. 
They utilise a number of the latest sequence learning
tricks such as scheduled sampling~\cite{bengio2015scheduled} and
attention~\cite{chorowski2015attention};
we take many inspirations from this work.


\section{Architecture}
\label{sec:arc}
\spacesection

\begin{figure*}[th!]
\centering 
\fbox{\includegraphics[width=1\textwidth]{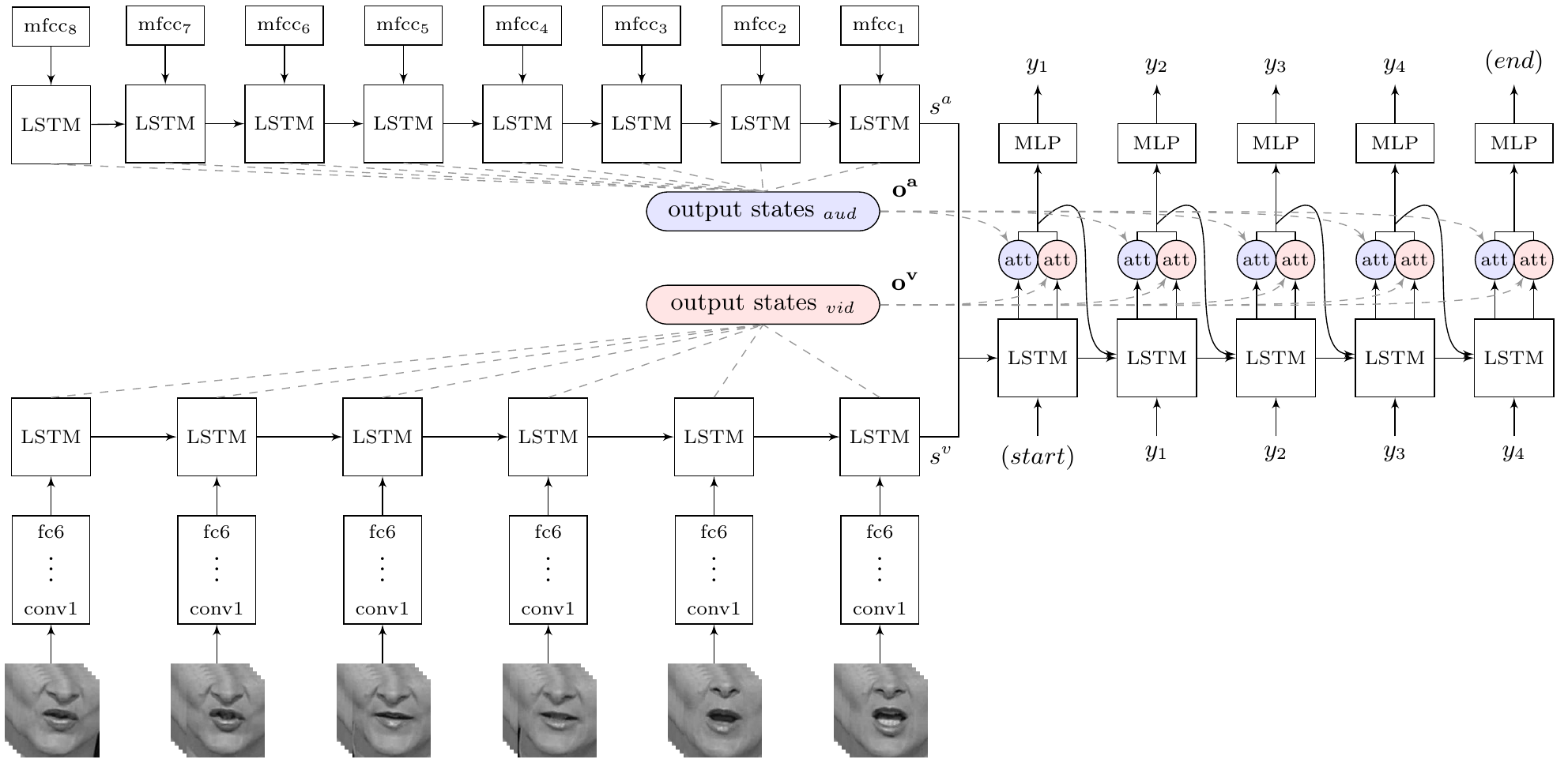}}
\vspace{-10pt}
\caption{{\it Watch, Listen, Attend and Spell} architecture. At each time step, the 
decoder outputs a character $y_i$, as well as two attention vectors. The attention vectors are used to select the appropriate period of the input visual
and audio sequences.}
\label{fig:arc} 
\vspace{-5pt}
\end{figure*}

In this section, we describe the 
{\it Watch, Listen, Attend and Spell} network that learns to predict characters
in sentences being spoken from a video of a talking face, with or without audio. 

We model each character $y_i$ in the output character sequence 
$\mathbf{y} = (y_1, y_2, ..., y_l)$ 
as a conditional distribution of the 
 previous characters $y_{<i}$, 
 the input image sequence $\mathbf{x}^v = (x^v_1, x^v_2, ..., x^v_n)$
for lip reading, and
 the input audio sequence $\mathbf{x}^a = (x^a_1, x^a_2, ..., x^a_m)$. 
Hence, we model the output probability distribution as:

\spacepreeqn
\begin{align}
    P(\mathbf{y}|\mathbf{x}^v, \mathbf{x}^a) =  \prod_{i} P(y_i|\mathbf{x}^v,\mathbf{x}^a,y_{<i})
  \label{equ:condprob}
\end{align}
\spacepreeqn

Our model, which is summarised in Figure~\ref{fig:arc}, consists of three key
components: the image encoder $\mathtt{Watch}$ (Section~\ref{sec:net_watch}), the audio encoder $\mathtt{Listen}$ (Section~\ref{sec:net_listen}), and the character decoder $\mathtt{Spell}$ (Section~\ref{sec:net_spell}).
Each encoder transforms the respective input sequence into a fixed-dimensional
state vector $s$, and sequences of encoder outputs $\mathbf{o} = (o_1, ..., o_p)$,
$p \in (n,m)$; the decoder ingests the state and the attention vectors from
both encoders and produces a probability distribution over the output character sequence.

\spacepreeqn
\begin{align}
s^v, \mathbf{o}^v &=  \mathtt{Watch}(\mathbf{x}^v) \\
s^a, \mathbf{o}^a &=  \mathtt{Listen}(\mathbf{x}^a) \\
  P(\mathbf{y}|\mathbf{x}^v, \mathbf{x}^a) &=  \mathtt{Spell}(s^v,s^a,\mathbf{o}^v,\mathbf{o}^a)
\end{align}
\spacepreeqn

The three modules in the model are trained jointly. We describe the modules next, with implementation details given in Section~\ref{sec:implem}.

\subsection{Watch: Image encoder}
\label{sec:net_watch}
\spacesubsection

The image encoder consists of the convolutional module that generates image features ${f}^v_i$ for
every input timestep $x^v_i$, 
and the recurrent module that produces the fixed-dimensional state vector $s^v$ 
and a set of output vectors $\mathbf{o}^v$.

\spacepreeqn
\begin{align}
f^v_i &= \mathtt{CNN}(x^v_i) \\
h^v_i, o^v_i &= \mathtt{LSTM}(f^v_i,h^v_{i+1}) \label{eq:enclstm} \\
s^v &= h^v_1
\end{align}
\spacepreeqn

\begin{figure*}[ht!]
\centering 
\fbox{\includegraphics[width=1\textwidth]{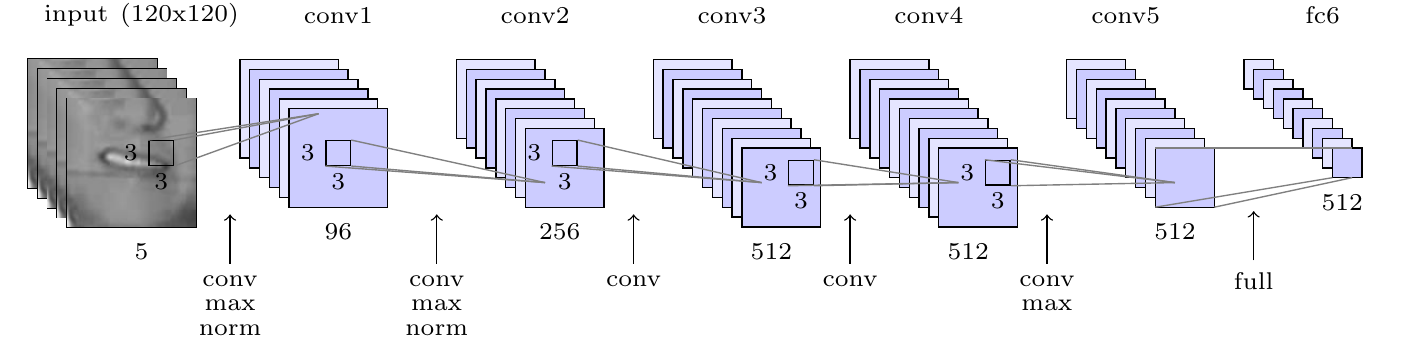}}
\vspace{-10pt}
\caption{The ConvNet architecture. The input is five gray level frames centered on the mouth region. The 512-dimensional fc6 vector forms the input to the LSTM.}
\label{fig:cnnarc} 
\vspace{-5pt}
\end{figure*}

The convolutional network is based on the VGG-M model~\cite{Chatfield14},
as it is memory-efficient, fast to train and has a decent 
classification performance on ImageNet~\cite{Russakovsky15}. 
The ConvNet layer configuration is shown in Figure~\ref{fig:cnnarc},
and is abbreviated as 
{\it conv1}  $\dotsm$ ${\it fc6}$ in the main network diagram.

The encoder LSTM network consumes the output features $f^v_i$ produced by the 
ConvNet at every input timestep, and generates a fixed-dimensional
state vector $s^v$. 
In addition, it produces an output vector $o^v_i$ at every timestep $i$. 
Note that the network ingests the inputs in reverse time order
(as in Equation~\ref{eq:enclstm}), 
which has shown to improve results in~\cite{Sutskever14}.

\subsection{Listen: Audio encoder}
\label{sec:net_listen}
\spacesubsection

The $\mathtt{Listen}$ module is an LSTM encoder similar to the $\mathtt{Watch}$ module,
without the convolutional part.
The LSTM directly ingests 13-dimensional MFCC features in reverse time order, and produces the state vector
$s^a$ and the output vectors $\mathbf{o}^a$.

\spacepreeqn
\begin{align}
h^a_j, o^a_j &= \mathtt{LSTM}(x^a_j,h^a_{j+1}) \\
s^a &= h^a_1
\end{align}
\spacepreeqn

\subsection{Spell: Character decoder}
\label{sec:net_spell}
\spacesubsection

The $\mathtt{Spell}$ module is based on a LSTM transducer~\cite{bahdanau2014neural,chorowski2015attention,chan2015listen}, here we add a dual attention mechanism.
At every output step $k$, the decoder LSTM produces the decoder states $h^d_k$ and output vectors $o^d_k$
from the previous step context vectors $c^v_{k-1}$ and $c^a_{k-1}$, output $y_{k-1}$
and decoder state $h^d_{k-1}$. 
The attention vectors are generated from 
the attention mechanisms
$\mathtt{Attention^v}$ and $\mathtt{Attention^a}$. 
The inner working of the attention mechanisms 
is described in \cite{bahdanau2014neural}, and repeated 
in the supplementary material.
We use two independent attention mechanisms for the lip and the audio 
input streams to refer to the asynchronous inputs with different sampling rates.
The attention vectors are fused with the output states (Equations~\ref{eq:attl} and \ref{eq:atta})
to produce the context vectors $c^v_k$ and $c^a_k$ that encapsulate the information 
required to produce the next step output. The probability distribution of the output
character is generated by an MLP with softmax over the output.

\spacepreeqn
\begin{align}
h^d_k, o^d_k &= \mathtt{LSTM}(h^d_{k-1},y_{k-1},c^v_{k-1},c^a_{k-1}) \label{eq:declstm}  \\
c^v_k &= \mathbf{o}^v \cdot \mathtt{Attention^v}(h^d_k,\mathbf{o}^v) \label{eq:attl} \\
c^a_k &= \mathbf{o}^a \cdot \mathtt{Attention^a}(h^d_k,\mathbf{o}^a)  \label{eq:atta} \\
  P(y_i|\mathbf{x}^v, \mathbf{x}^a, y_{<i}) &= \mathtt{softmax}(\mathtt{MLP}(o^d_k,c^v_k,c^a_k) )
\end{align}
\spacepreeqn

At $k=1$, the final encoder states 
$s_l$ and $s_a$ are used as the input instead of the previous decoder state
-- {\em i.e.} $h^d_0 = \mathtt{concat}(s^a,s^v)$ -- to 
help produce the context vectors $c^v_1$ and $c^a_1$ in the absence of the previous state or context.

\noindent{\bf Discussion.} 
In our experiments, we have observed that the attention mechanism
is absolutely critical for the audio-visual speech recognition
system to work.
Without attention, the model appears to `forget' the input signal, 
and produces an output sequence that correlates very little to the
input, beyond the first word or two 
(which the model gets correct, as these are the last words to be seen
by the encoder). 
The attention-less model yields Word Error Rates over 100\%, so
we do not report these results.

The dual-attention mechanism allows the model to extract 
information from both audio and video inputs, even when one
stream is absent, or the two streams are not time-aligned.
The benefits are clear in the experiments with noisy or no audio
(Section~\ref{sec:exp}).

Bidirectional LSTMs have been used in many sequence learning
tasks~\cite{chan2015listen,graves2013hybrid,chorowski2015attention}
for its ability to produce outputs conditioned on 
future context as well as past context. 
We have tried replacing the unidirectional encoders in the 
$\mathtt{Watch}$ and $\mathtt{Listen}$ modules with
bidirectional encoders, 
however these networks took significantly longer to train, whilst
 providing no obvious performance improvement.
This is presumably because the $\mathtt{Decoder}$ module 
is anyway conditioned on the full
input sequence, so bidirectional encoders are not necessary for providing 
context, and the
attention mechanism suffices to provide the additional local focus.


\section{Training strategy}
\label{sec:training}
\spacesection

In this section, we describe the strategy used to effectively train the
{\it Watch, Listen, Attend and Spell} network, making best
use of the limited amount of data available.

\subsection{Curriculum learning} 
\spacesubsection

Our baseline strategy is to train the model from scratch, 
using the full sentences from the `Lip Reading Sentences' dataset --
previous works in speech recognition have
taken this approach. 
However, as \cite{chan2015listen} reports, 
the LSTM network converges very slowly when the number
of timesteps is large, because the decoder initially
has a hard time extracting the relevant information
 from all the input steps.

We introduce a new strategy where we start training only on single
word examples, and then let the sequence length grow as the 
network trains. 
These short sequences are parts of the longer sentences in the dataset.
We observe that the rate of convergence on the training set 
is several times faster, 
and it also significantly reduces overfitting, presumably because it 
works as a natural way of augmenting the data.
The test performance improves by a large margin, reported in 
Section~\ref{sec:exp}.

\subsection{Scheduled sampling}
\spacesubsection

When training a recurrent neural network, one typically uses the
previous time step ground truth as the next time step input, which
helps the model learn a kind of language model over target
tokens. However during inference, the previous step ground-truth is
unavailable, resulting in poorer performance because the model was not
trained to be tolerant to feeding in bad predictions at some time
steps. We use the scheduled sampling method of Bengio {\em et
al.}~\cite{bengio2015scheduled} to bridge this discrepancy between how
the model is used at training and inference. At train time, we
randomly sample from the previous output, instead of always 
using the ground-truth.  When training on shorter sub-sequences,
ground-truth previous characters are used.  When training on full
sentences, the sampling probability from the previous output was
increased in steps from 0 to 0.25 over time. We were not able to
achieve stable learning at sampling probabilities of greater than
0.25.

\subsection{Multi-modal training}
\label{sec:multimodal}
\spacesubsection

Networks with multi-modal inputs can often be dominated by one of the modes \cite{Feichtenhofer16}.
In our case we observe that the audio signal dominates, because 
speech recognition is a significantly
easier problem than lip reading. 
To help prevent this from happening,
one of the following input types is uniformly
selected at train time for each example:
(1) audio only;
(2) lips only;
(3) audio and lips.

If mode (1) is selected, the audio-only data described in
Section~\ref{sec:audiodata} is used. 
Otherwise, the standard audio-visual data is used.

We have over 300,000 sentences in the recorded data, but only
around 100,000 have corresponding facetracks. 
In machine translation, it has been shown that monolingual dummy data
can be used to help improve the performance of a translation model~\cite{sennrich2015improving}.
By similar rationale, we use the sentences without facetracks as supplementary training data
to boost audio recognition performance and to build a richer language model to help
improve generalisation.

\subsection{Training with noisy audio}
\label{sec:trainwithnoise}
\spacesubsection

The WLAS model is initially 
trained with clean input audio for
faster convergence.
To improve the model's tolerance to audio noise, 
we apply additive white Gaussian noise with SNR of 10dB (10:1 ratio of the signal power to the noise power) and 0dB (1:1 ratio)
later in training.

\subsection{Implementation details}
\label{sec:implem}
\spacesubsection
The input images are 120$\times$120 in dimension, 
and are sampled at 25Hz. 
The image only covers the lip region of the face, as shown in Figure~\ref{fig:pm}. 
The ConvNet ingests 5-frame sliding windows
using the Early Fusion method of ~\cite{Chung16}, 
moving 1-frame at a time.
The MFCC features are calculated over 25ms 
windows and at 100Hz,
with a time-stride of 1.
For $\mathtt{Watch}$ and $\mathtt{Listen}$
modules, we use a three layer LSTM with
cell size of 256. 
For the $\mathtt{Spell}$ module, we use a three
layer LSTM with cell size of 512.
The output size of the network is 45, for
every character in the alphabet, numbers,
common punctuations, and tokens for \texttt{[sos]}, \texttt{[eos]},
\texttt{[pad]}. The full list is given in the supplementary material.

Our implementation is based on the TensorFlow library~\cite{abadi2016tensorflow}
and trained on a GeForce Titan X GPU with 12GB memory. 
The network is trained 
using stochastic gradient descent with a batch size of 64
and with dropout and label smoothing.
The layer weights of the convolutional layers are initialised from the visual stream of~\cite{Chung16a}. 
All other weights are randomly initialised.
 
An initial learning rate of 0.1 was used, and decreased by
10\% every time the training error did not improve for 2,000 iterations. 
Training on the full sentence data was stopped when the validation error
 did not improve for 5,000 iterations. 
The model was trained for around 500,000 iterations, 
which took approximately 10 days.


\section{Dataset}
\label{sec:dataset}
\spacesection

\begin{figure*}[th!]
\centering 

\includegraphics[width=.246\textwidth]{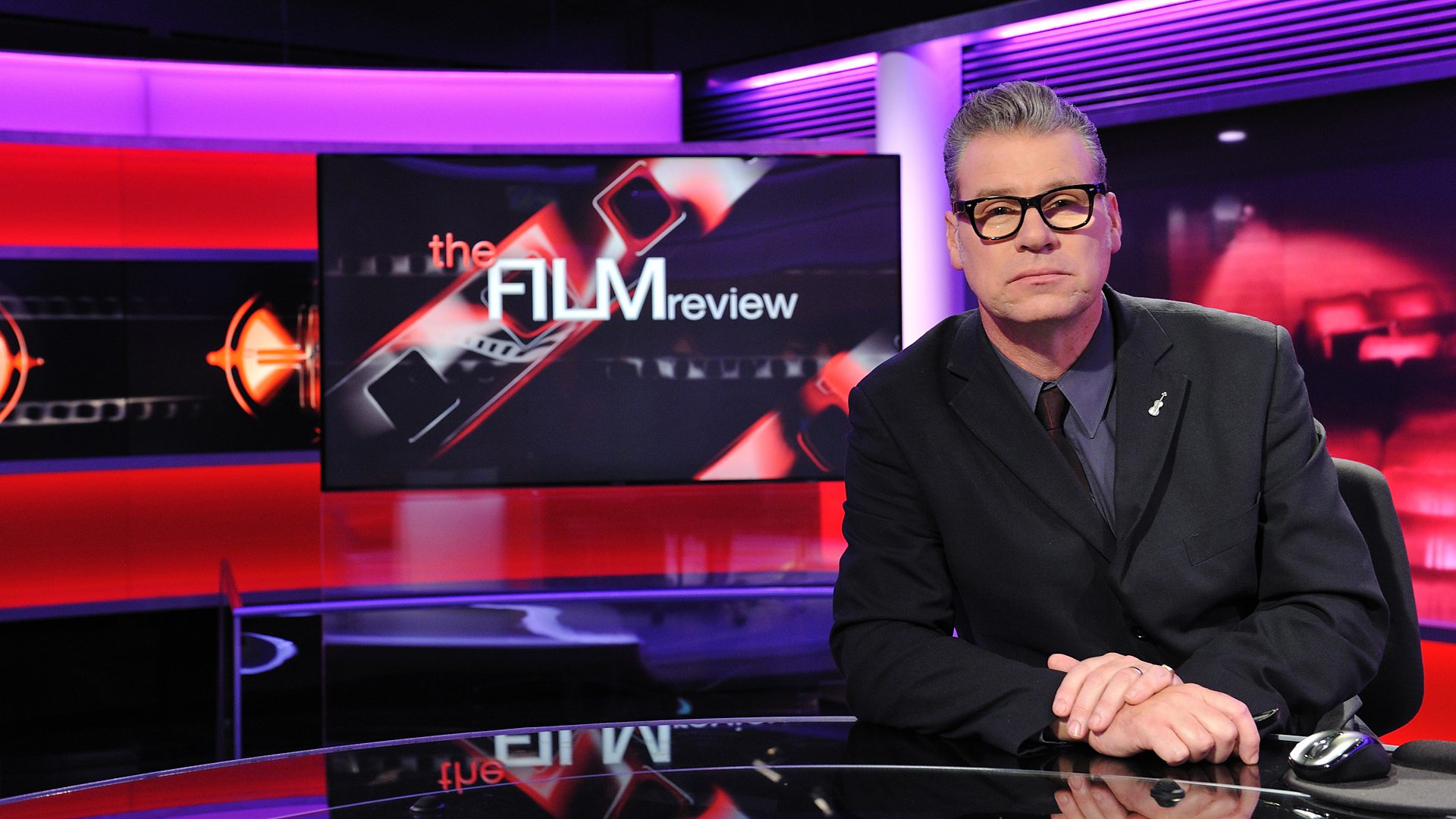} 
\includegraphics[width=.246\textwidth]{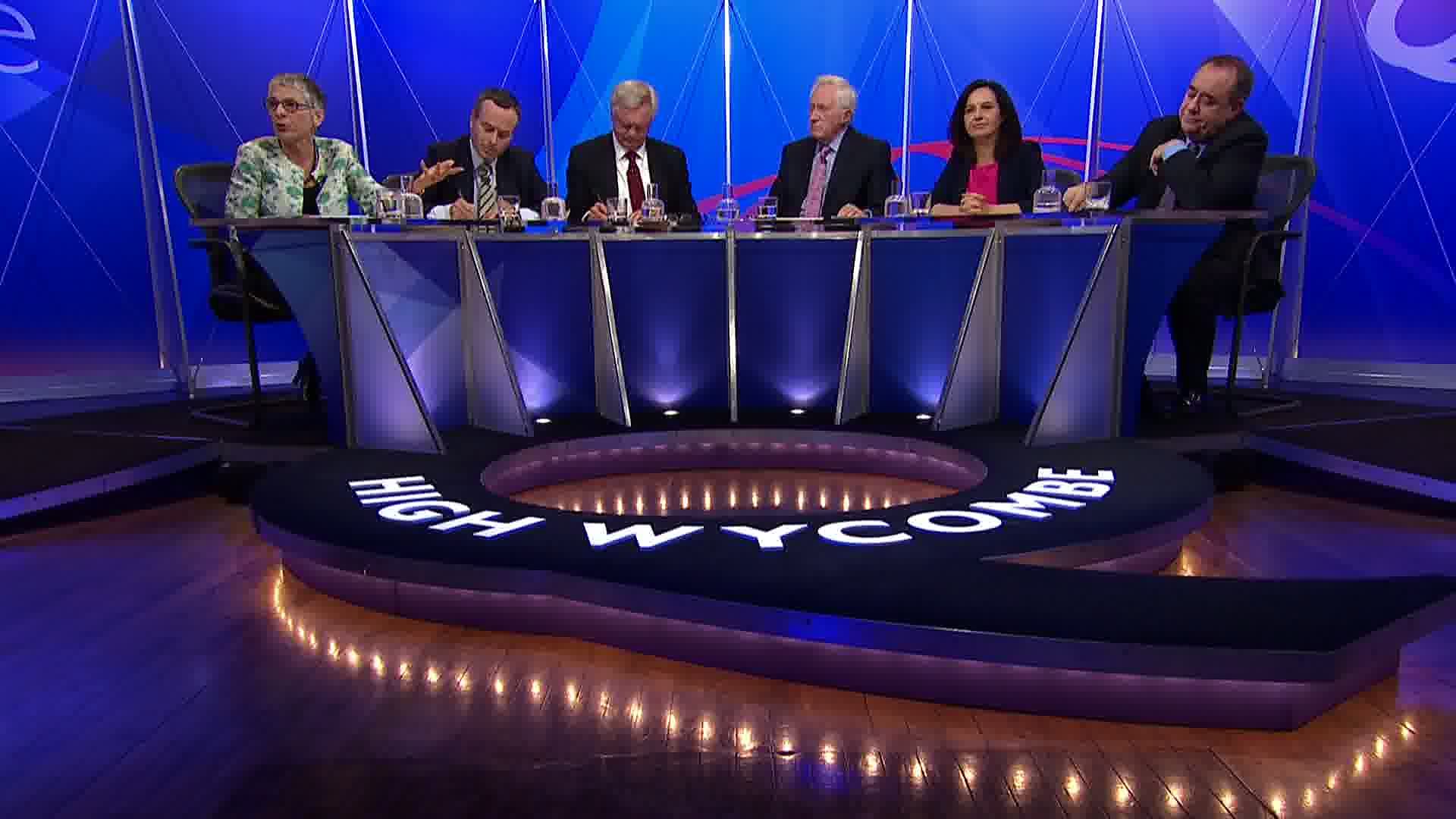}
\includegraphics[width=.246\textwidth]{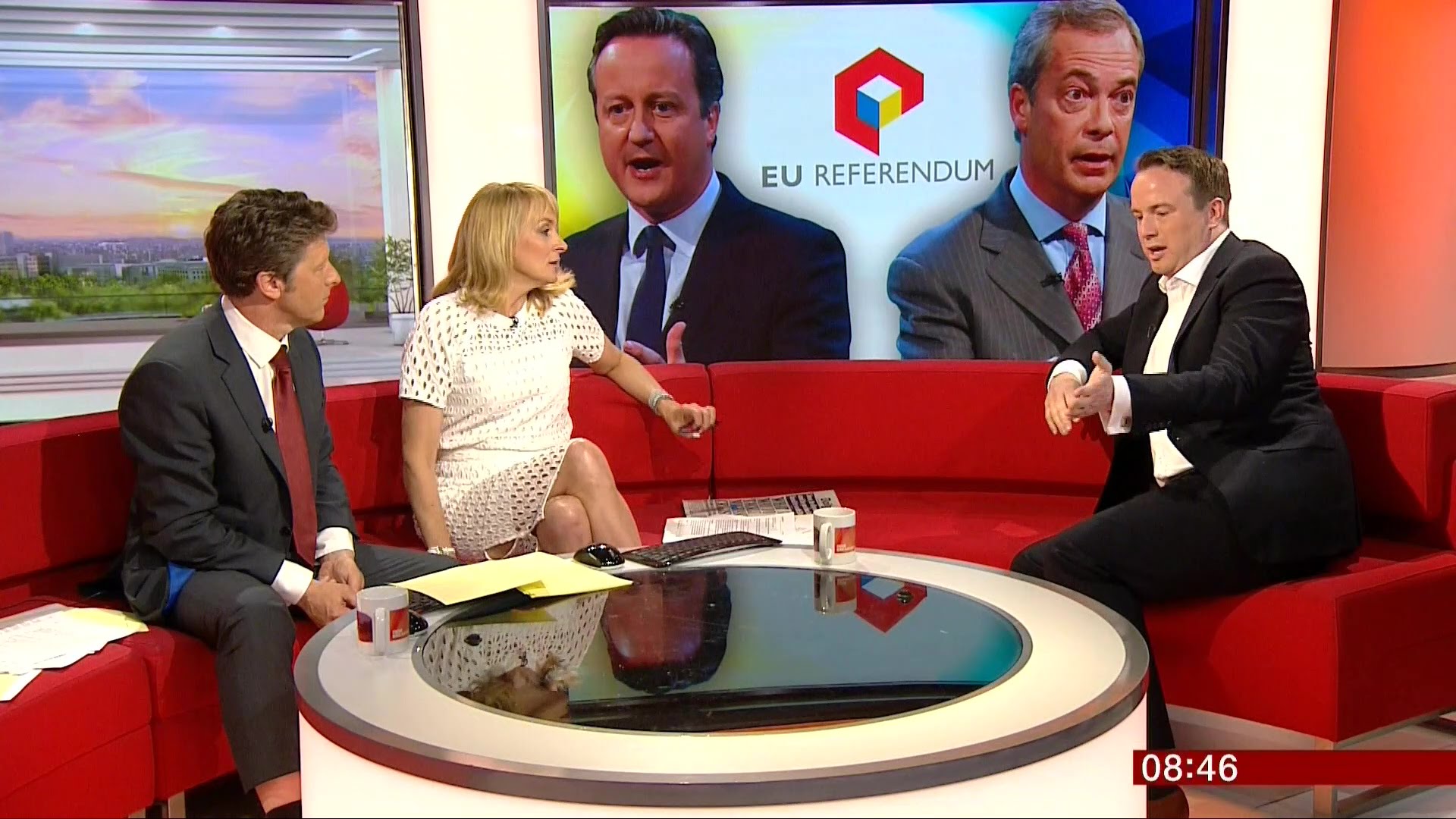}
\includegraphics[width=.246\textwidth]{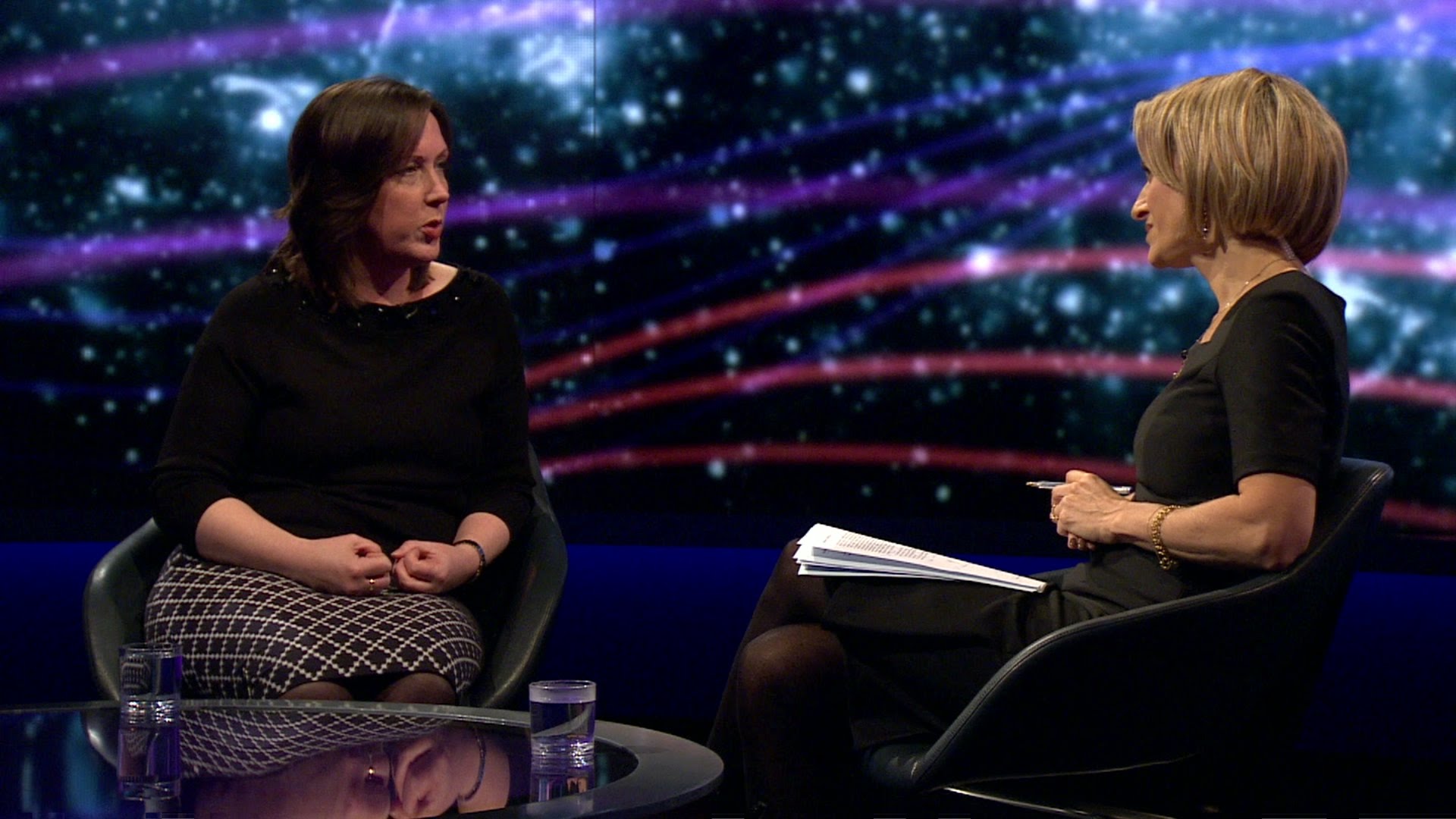} 

\vspace{5pt}

\includegraphics[width=0.058\textwidth]{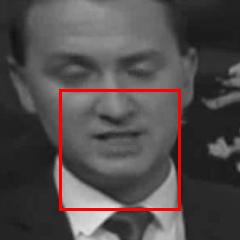}
\includegraphics[width=0.058\textwidth]{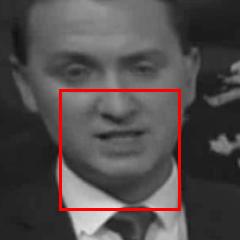}
\includegraphics[width=0.058\textwidth]{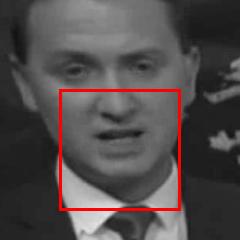}
\includegraphics[width=0.058\textwidth]{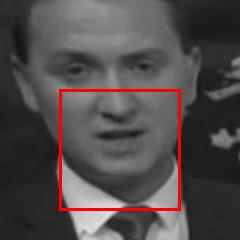}
\includegraphics[width=0.058\textwidth]{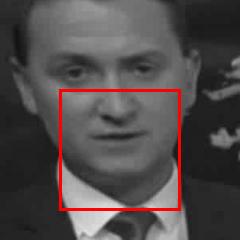}
\includegraphics[width=0.058\textwidth]{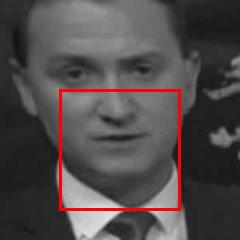}
\includegraphics[width=0.058\textwidth]{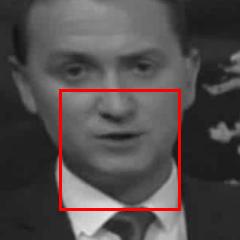}
\includegraphics[width=0.058\textwidth]{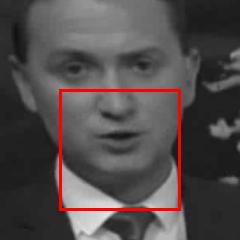} 
\includegraphics[width=0.058\textwidth]{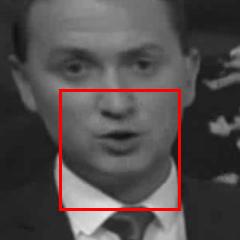}
\includegraphics[width=0.058\textwidth]{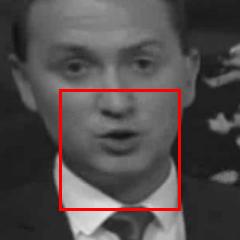}
\includegraphics[width=0.058\textwidth]{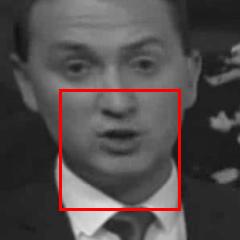}
\includegraphics[width=0.058\textwidth]{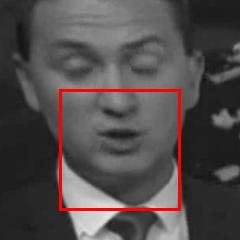}
\includegraphics[width=0.058\textwidth]{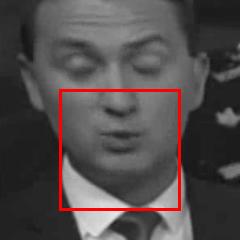}
\includegraphics[width=0.058\textwidth]{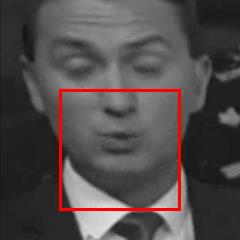}
\includegraphics[width=0.058\textwidth]{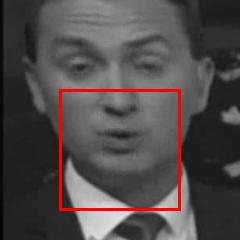}
\includegraphics[width=0.058\textwidth]{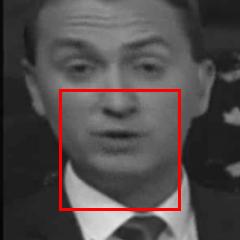}

\vspace{5pt}

\includegraphics[width=0.058\textwidth]{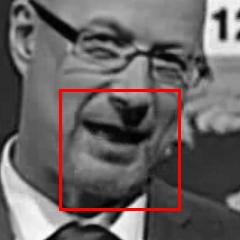}
\includegraphics[width=0.058\textwidth]{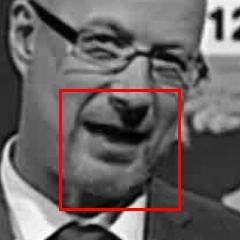}
\includegraphics[width=0.058\textwidth]{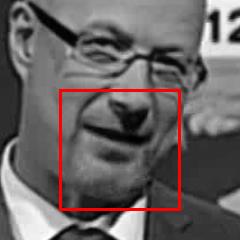}
\includegraphics[width=0.058\textwidth]{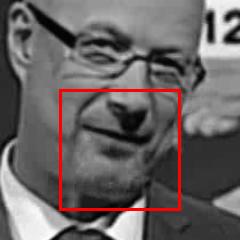}
\includegraphics[width=0.058\textwidth]{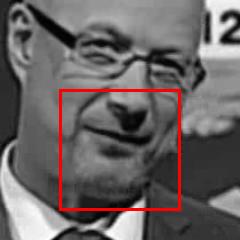}
\includegraphics[width=0.058\textwidth]{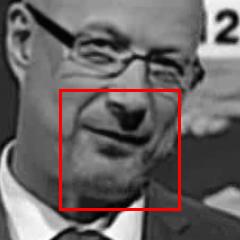}
\includegraphics[width=0.058\textwidth]{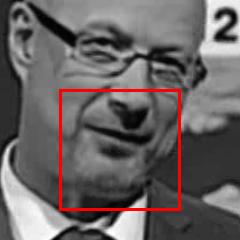}
\includegraphics[width=0.058\textwidth]{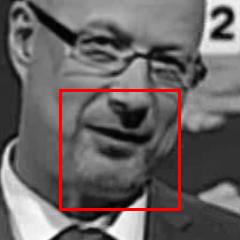} 
\includegraphics[width=0.058\textwidth]{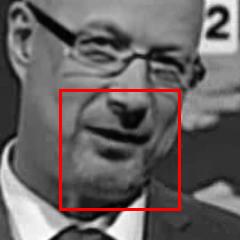}
\includegraphics[width=0.058\textwidth]{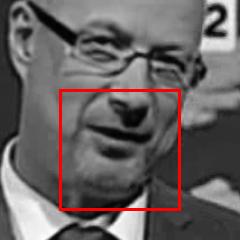}
\includegraphics[width=0.058\textwidth]{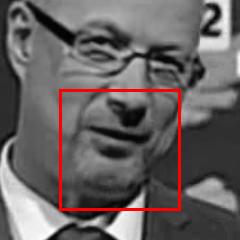}
\includegraphics[width=0.058\textwidth]{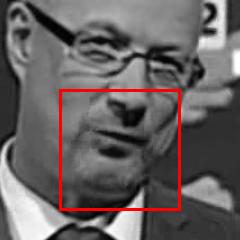}
\includegraphics[width=0.058\textwidth]{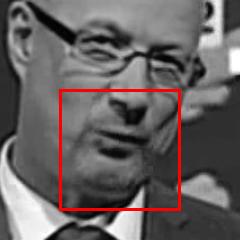}
\includegraphics[width=0.058\textwidth]{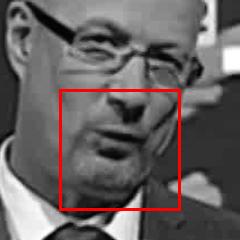}
\includegraphics[width=0.058\textwidth]{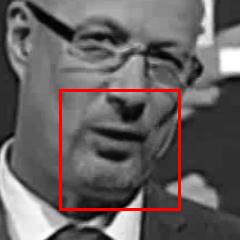}
\includegraphics[width=0.058\textwidth]{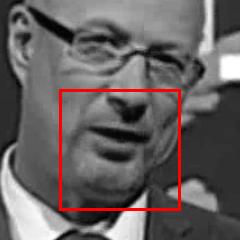}

\caption{{\bf Top:} Original still images from the BBC lip reading dataset -- 
 News, Question Time, Breakfast, Newsnight (from left to right).
 {\bf Bottom:} The mouth motions for `afternoon' from two different speakers.
The network sees the areas inside the red squares.}
\label{fig:pm} 
\vspace{-5pt}
\end{figure*}

In this section, we describe the multi-stage pipeline for 
automatically generating a large-scale dataset for
audio-visual speech recognition. 
Using this pipeline, we have been
able to collect thousands of hours of spoken sentences and
phrases along with the corresponding facetrack.
We use a variety of BBC programs
recorded between 2010 and 2016,
listed in Table~\ref{table:programs},
and shown in Figure~\ref{fig:pm}.

The selection of programs are deliberately 
similar to those used by \cite{Chung16} for 
two reasons: 
(1) a wide range of speakers appear
in the news and the debate programs, unlike dramas
with a fixed cast; 
(2) shot changes are less frequent,
therefore there are more full sentences with 
continuous facetracks.

\begin{table}[ht]
\centering
\begin{tabular}{| l | l | r | r | }
  \hline
  \textbf{Channel} & \textbf{Series name} & \textbf{\# hours} & \textbf{\# sent.} \\ \hline \hline 
 BBC 1 HD & News$^\dag$ & 1,584 & 50,493\\ \hline 
 BBC 1 HD & Breakfast & 1,997 & 29,862 \\ \hline 
 BBC 1 HD & Newsnight     &  590 & 17,004  \\ \hline 
 BBC 2 HD & World News  & 194 & 3,504 \\ \hline 
 BBC 2 HD & Question Time & 323 & 11,695 \\ \hline
 BBC 4 HD & World Today & 272 & 5,558 \\ \hline \hline
 \textbf{All} &  & 4,960  & 118,116 \\ \hline
\end{tabular} 
\spacetablecapt
\caption{Video statistics. 
The number of hours of the original BBC video;
the number of sentences with full facetrack. 
$^\dag$BBC News at 1, 6 and 10.  }
\label{table:programs}
\vspace{-5pt}
\end{table}

\begin{figure}[ht]
\centering 
\includegraphics[width=1\linewidth]{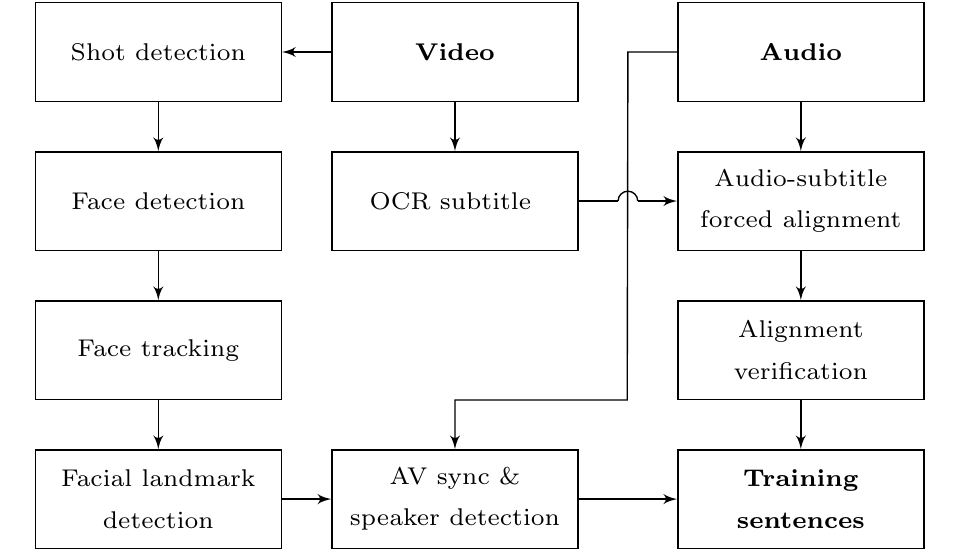}
\vspace{-10pt}
\caption{Pipeline to generate the dataset.}
\label{fig:pipeline} 
\vspace{-5pt}
\end{figure}

The processing pipeline is summarised in Figure~\ref{fig:pipeline}. 
Most of the steps are based on the methods
described in \cite{Chung16} and \cite{Chung16a},
but we give a brief sketch of the method here.

\noindent\textbf{Video preparation.} 
First, shot boundaries are detected by comparing colour histograms
across consecutive frames~\cite{Lienhart01}.
The HOG-based face detection~\cite{king2009dlib} is then
performed on every frame of the video.
The face detections of the same person are grouped 
across frames using a KLT tracker \cite{Tomasi92a}. 
Facial landmarks are extracted from a sparse 
subset of pixel intensities
using an ensemble of regression
trees \cite{kazemi2014one}.

\noindent\textbf{Audio and text preparation.}
The subtitles in BBC videos are not broadcast
in sync with the audio.
The Penn Phonetics Lab Forced
Aligner~\cite{yuan2008speaker,hermansky1990perceptual} 
is used to
force-align the subtitle to the audio signal.
Errors exist in the alignment as the transcript is
not verbatim --
therefore the aligned labels are filtered by
checking against the commercial IBM Watson Speech to Text
service.

\noindent\textbf{AV sync and speaker detection.}
In BBC videos, the audio and the video streams can be out of
sync by up to around one second, which can cause problems when
the facetrack corresponding to a sentence is being extracted.
The two-stream network described in \cite{Chung16a} is used to 
synchronise the two streams. 
The same network is also used to determine who is speaking
in the video, and reject the clip if it is a voice-over.

\noindent\textbf{Sentence extraction.} 
The videos are divided
into invididual sentences/ phrases using the punctuations
in the transcript. 
The sentences are separated by full stops, commas 
and question marks; and
are clipped to 100 characters or 10 
seconds, due to GPU memory constraints.
We do not impose any restrictions on the vocabulary size.

The training, validation 
and test sets are divided according to broadcast date, 
and the dates of 
videos corresponding to each set 
are shown in Table~\ref{table:datastat}. The dataset contains thousands of
different speakers which enables the model to be speaker agnostic.

\begin{table}[th!]
\centering
\begin{tabular}{| l | c | r | r |  }
  \hline \textbf{Set} & \textbf{Dates} & \textbf{\# Utter.} & \textbf{Vocab}  \\ 
  \hline \hline 

    Train  & 01/2010 - 12/2015   & 101,195 & 16,501  \\ \hline 
    Val   & 01/2016 - 02/2016   & 5,138 & 4,572  \\ \hline 
    Test  & 03/2016 - 09/2016   & 11,783 &  6,882  \\ \hline 
     \hline
    \textbf{All}  &    & 118,116 &  17,428  \\ \hline 
\end{tabular} 
\spacetablecapt
\caption{{\bf The Lip Reading Sentences (LRS) audio-visual dataset.} Division of training, validation and test data; and the number of utterances and vocabulary size of each partition.
Of the 6,882 words in the test set, 6,253 are in the
training or the validation sets; 
6,641 are in the audio-only training data.
{\bf Utter:} Utterances
}
\label{table:datastat}
\vspace{-5pt}
\end{table}

Table~\ref{table:existingdata} compares the {\it `Lip Reading Sentences'} (LRS)
dataset to the largest existing public datasets.

\begin{table}[ht]

\centering
\begin{tabular}{| l | l |  r | r | r | }
  \hline
  \textbf{Name}                            & \textbf{Type} &  \textbf{Vocab} 
   & \textbf{\# Utter.} & \textbf{\# Words} \\ \hline 

 GRID   \cite{cooke2006audio}              & Sent. & 51         & 33,000 & 165,000 \\ \hline 
 LRW  \cite{Chung16}                       & Words   & 500        & 400,000 & 400,000 \\ \hline 
 {\bf LRS}                                 & Sent.   & 17,428    & 118,116 & 807,375 \\ \hline 
\end{tabular} 
\spacetablecapt
\caption{
Comparison to existing large-scale lip reading datasets.
}
\label{table:existingdata}
\vspace{-5pt}
\end{table}

\subsection{Audio-only data}
\label{sec:audiodata}
\spacesubsection

In addition to the audio-visual dataset, we prepare an auxiliary
audio-only {\em training} dataset.  These are the sentences in the BBC
programs for which facetracks are not available.  The use of this data
is described in Section~\ref{sec:multimodal}. It is only used for
training, not for testing.

\begin{table}[th!]
\centering
\begin{tabular}{| l | c | r | r | }
  \hline \textbf{Set} & \textbf{Dates} & \textbf{\# Utter.} & \textbf{Vocab}  \\ 
  \hline \hline 

    Train  & 01/2010 - 12/2015   & 342,644 & 25,684  \\ \hline 
\end{tabular} 
\spacetablecapt
\caption{Statistics of the Audio-only training set.}
\label{table:audiodatastat}
\vspace{-10pt}
\end{table}


\section{Experiments}
\label{sec:exp}
\spacesection

In this section we evaluate and compare the proposed architecture
and training strategies. 
We also compare our method
to the state of the art on public
benchmark datasets.

To clarify which of the modalities are being used, 
we call the models in lips-only and audio-only experiments
{\em Watch, Attend and Spell} (WAS),  {\em Listen, Attend and Spell} (LAS) respectively.
These are the same {\em Watch, Listen, Attend and Spell} model with
either of the inputs disconnected and replaced with all-zeros.

\subsection{Evaluation.}
\label{sec:eval}
\spacesubsection

The models are trained on the LRS dataset (the train/val partition) and 
the Audio-only training dataset
(Section~\ref{sec:dataset}). 
The inference and evaluation procedures are 
described below.

\begin{figure}[ht]
\centering 
\includegraphics[width=1\linewidth]{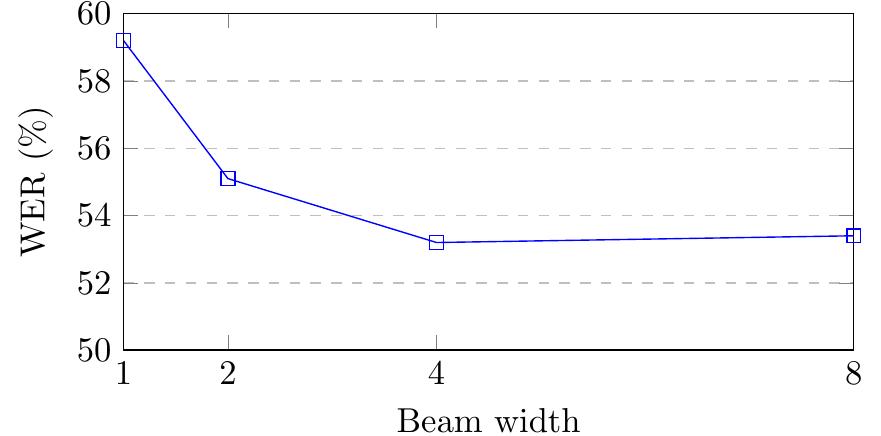}
\vspace{-10pt}
\caption{The effect of beam width on Word Error Rate.}
\label{fig:beam} 
\vspace{-5pt}
\end{figure}

\noindent{\bf Beam search. } 
Decoding is performed with beam search of width 4, 
in a similar manner to~\cite{Sutskever14,chan2015listen}.
At each timestep, the hypotheses in the beam are
expanded with every possible character, and only
the 4 most probable hypotheses are stored.
Figure~\ref{fig:beam} shows the effect of increasing
the beam width -- there is no observed benefit for
increasing the width beyond 4.

\noindent{\bf Evaluation protocol. } 
The models are evaluated on an independent test set
(Section~\ref{sec:dataset}). 
For all experiments,
we report the Character Error Rate (CER), 
the Word Error Rate (WER)
and the BLEU metric.
CER and WER are defined as
$\mathtt{Error Rate} = (S+D+I)/N$, 
where $S$ is the number of substitutions,
$D$ is the number of deletions,
$I$ is the number of insertions
to get from the reference to the hypothesis,
and $N$ is the number of words in the reference.
BLEU~\cite{papineni2002bleu} is a modified form
of n-gram precision to compare a candidate sentence
to one or more reference sentences.
Here, we use the unigram BLEU.

\begin{table}[th!]
\centering
\begin{tabular}{| l | r | r | r | r | }
  \hline \textbf{Method} & \textbf{SNR} & \textbf{CER} & \textbf{WER} &\textbf{BLEU$^\dag$}\\ \hline \hline 

    \multicolumn{5}{|c|}{{\bf Lips only}} \\ \hline
    Professional$^\ddag$   & - & 58.7\%    & 73.8\%  & 23.8 \\ \hline 
    WAS                     & - & 59.9\%    & 76.5\%  & 35.6 \\ \hline 
    WAS+CL                  & - & 47.1\%    & 61.1\%  & 46.9 \\ \hline 
    WAS+CL+SS               & - & 42.4\%    & 58.1\%  & 50.0\\ \hline
    WAS+CL+SS+BS            & - & 39.5\%    & 50.2\%  & 54.9\\ \hline \hline 

    \multicolumn{5}{|c|}{{\bf Audio only}} \\ \hline
    Google Speech API & clean   & 17.6\% & 22.6\% & 78.4 \\ \hline
    Kaldi SGMM+MMI$^\star$& clean   &  9.7\% & 16.8\% & 83.6 \\ \hline
    LAS+CL+SS+BS      & clean   & 10.4\% & 17.7\% & 84.0 \\ \hline
    LAS+CL+SS+BS      & 10dB    & 26.2\% & 37.6\% & 66.4\\ \hline
    LAS+CL+SS+BS      & 0dB     & 50.3\% & 62.9\% & 44.6\\ \hline
     \hline

    \multicolumn{5}{|c|}{{\bf Audio and lips}} \\ \hline
    WLAS+CL+SS+BS & clean   & 7.9\% & 13.9\% & 87.4 \\ \hline
    WLAS+CL+SS+BS & 10dB    & 17.6\% & 27.6\% & 75.3  \\ \hline
    WLAS+CL+SS+BS & 0dB     & 29.8\% & 42.0\% & 63.1\\ \hline

\end{tabular} 
\spacetablecapt
\caption{
Performance on the LRS test set. 
{\bf WAS}: {\it Watch, Attend and Spell}; 
{\bf LAS}: {\it Listen, Attend and Spell}; 
{\bf WLAS}: {\it Watch, Listen, Attend and Spell}; 
{\bf CL}: Curriculum Learning;
{\bf SS}: Scheduled Sampling;
{\bf BS}: Beam Search.
 $\dag$Unigram BLEU with brevity penalty.
 $\ddag$Excluding samples that the lip reader
 declined to annotate. Including these, the
 CER rises to 78.9\% and the WER to 87.6\%. 
 $\star$~The
Kaldi SGMM+MMI model used here achieves a WER of 3.6\% on the WSJ (eval92)
test set, which is within 0.2\% of the current state-of-the-art. The
acoustic and language models have been re-trained on our dataset.
 }
\label{table:mainresults}
\vspace{-5pt}
\end{table}

\noindent\textbf{Results.}
All of the training methods discussed in 
Section~\ref{sec:training} contribute to
improving the performance. 
A breakdown of this is given in Table~\ref{table:mainresults}
for the lips-only experiment.
For all other experiments, we only report results
obtained using the best strategy.

\noindent\textbf{Lips-only examples.}
The model learns to correctly
predict extremely complex unseen
sentences from a wide range of content --
examples are shown in Table~\ref{table:lipres}.

\begin{table}[ht]
\centering
\footnotesize
\begin{tabular}{| p{7cm} |  }
  \hline
  MANY MORE PEOPLE WHO WERE INVOLVED IN THE ATTACKS \\ \hline 
  CLOSE TO THE EUROPEAN COMMISSION'S MAIN BUILDING  \\ \hline 
  WEST WALES AND THE SOUTH WEST AS WELL AS WESTERN SCOTLAND \\ \hline 
  WE KNOW THERE WILL BE HUNDREDS OF JOURNALISTS HERE AS WELL \\ \hline 
  ACCORDING TO PROVISIONAL FIGURES FROM THE ELECTORAL COMMISSION \\ \hline 
  THAT'S THE LOWEST FIGURE FOR EIGHT YEARS \\ \hline 
  MANCHESTER FOOTBALL CORRESPONDENT FOR THE DAILY MIRROR \\ \hline 
  LAYING THE GROUNDS FOR A POSSIBLE SECOND REFERENDUM \\ \hline 
  ACCORDING TO THE LATEST FIGURES FROM THE OFFICE FOR NATIONAL STATISTICS \\ \hline 
  IT COMES AFTER A DAMNING REPORT BY THE HEALTH WATCHDOG \\ \hline

\end{tabular} 
\normalsize
\spacetablecapt
\caption{Examples of unseen sentences that WAS correctly predicts
(lips only).}
\label{table:lipres}
\vspace{-5pt}
\end{table}

\noindent\textbf{Audio-visual examples.}
As we hypothesised, the results in 
(Table~\ref{table:mainresults}) demonstrate that
the mouth movements provide important
cues in speech recognition when the audio
signal is noisy; and also give an
improvement in performance 
even when the audio signal is clean -- the character error rate is
reduced from 10.4\% for audio only to 7.9\% for audio together lip reading.
Table~\ref{table:avres} shows some of
the many examples where the WLAS model fails to
predict the correct sentence from the lips or the audio
alone, but successfully deciphers the words when 
both streams are present.

\begin{table}[ht]
\centering
\footnotesize
\begin{tabular}{| p{1cm} | p{5.6cm} | }
  \hline 
 \textbf{GT} & IT WILL BE THE CONSUMERS \\ \hline 
 \textbf{A} & IN WILL BE THE CONSUMERS \\ \hline 
 \textbf{L} & IT WILL BE IN THE CONSUMERS \\ \hline 
 \textbf{AV} & IT WILL BE THE CONSUMERS \\ \hline \hline

 \textbf{GT} & CHILDREN IN EDINBURGH \\ \hline 
 \textbf{A} & CHILDREN AND EDINBURGH \\ \hline 
 \textbf{L} & CHILDREN AND HANDED BROKE \\ \hline 
 \textbf{AV} & CHILDREN IN EDINBURGH \\ \hline \hline

  \textbf{GT} & JUSTICE AND EVERYTHING ELSE \\ \hline 
 \textbf{A} & JUST GETTING EVERYTHING ELSE\\ \hline 
 \textbf{L} & CHINESES AND EVERYTHING ELSE \\ \hline 
 \textbf{AV} & JUSTICE AND EVERYTHING ELSE \\ \hline

\end{tabular} 
\normalsize
\spacetablecapt
\caption{Examples of AVSR results. 
{\bf GT:} Ground Truth;
{\bf A:} Audio only (10dB SNR);
{\bf L:} Lips only;
{\bf AV:} Audio-visual.
}
\label{table:avres}
\vspace{-5pt}
\end{table}

\begin{figure}[ht]
\centering 
\includegraphics[width=1\linewidth]{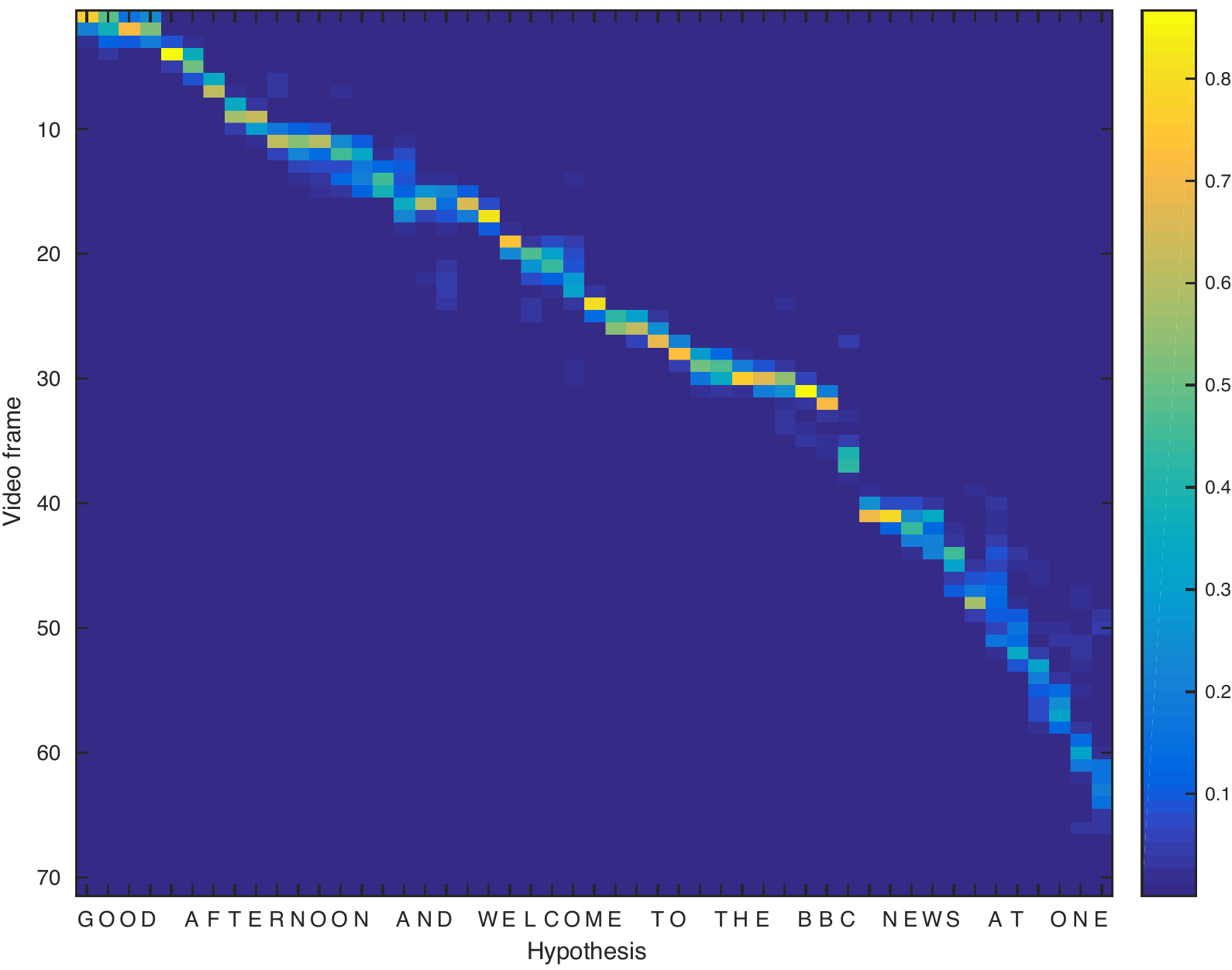}
\caption{Alignment between the video frames
and the character output.}
\label{fig:attn} 
\vspace{-5pt}
\end{figure}

\noindent{\bf Attention visualisation.}
The attention mechanism generates explicit alignment
between the input video frames (or the audio signal) 
and the hypothesised character output. 
Figure~\ref{fig:attn} visualises the alignment of the 
characters ``Good afternoon and welcome to the BBC News at One''
and the corresponding video frames.
{\em This result is better shown as a video;
please see supplementary materials.}

\noindent{\bf Decoding speed. }  
The decoding happens
significantly faster than real-time.
The model takes approximately 0.5 seconds to read and decode a
5-second sentence when using a beam width of 4.

\subsection{Human experiment}
\spacesubsection

In order to compare the performance of our model
to what a human can achieve, we instructed
a professional lip reading 
company
to decipher a random sample of 200 videos 
from our test set. 
The lip reader has around 10 years of professional
experience and deciphered videos in a range
of settings, e.g.\ forensic lip reading 
for use in court, the royal wedding, etc.

The lip reader was allowed to see
the full face (the whole picture in
the bottom two rows of
Figure~\ref{fig:pm}),
but not the background,
in order to prevent them from reading
subtitles or guessing the words
from the video content. However,
they were informed which program the 
video comes from, and were allowed to 
look at some videos from the training set with 
ground truth.

The lip reader was given 10 times the video
duration to predict the words being spoken,
and within that time, they were allowed to watch
the video as many times as they wished. Each of
the test sentences was up to 100 characters in length.

We observed that the professional lip reader is able
to correctly decipher less than one-quarter of the 
spoken words (Table~\ref{table:mainresults}).
This is consistent with previous studies on
the accuracy of human lip reading~\cite{marschark2010oxford}.
In contrast, the WAS model
(lips only) is able to decipher 
 half of the spoken words. 
Thus, this is significantly better than 
professional lip readers can achieve.

\subsection{LRW dataset}
\spacesubsection

The `Lip Reading in the Wild' (LRW) dataset consists of
up to 1000 utterances of 500 isolated words from 
BBC television, spoken
by over a thousand different speakers. 

\noindent{\bf Evaluation protocol.\ } 
The train, validation and test splits
are provided with the dataset.
We give word error rates.

\noindent{\bf Results.\ } The network is fine-tuned for one epoch to classify only the 500 word classes of this dataset's lexicon.
As shown in Table~\ref{table:extres},
our result exceeds the current
state-of-the-art on this dataset by a large margin.

\begin{table}[th!]
\centering
\begin{tabular}{| l | c | c | }
  \hline \textbf{Methods} & \textbf{LRW~\cite{Chung16}} 
  & \textbf{GRID~\cite{cooke2006audio}} \\ \hline \hline 


    Lan {\em et al.}~\cite{lan2009comparing}     & -   & 35.0\% \\ \hline 
    Wand {\em et al.}~\cite{wand2016lipreading}  & -   & 20.4\% \\ \hline 
    Assael {\em et al.}~\cite{yannis2016lipnet}  & -   & 4.8\% \\ \hline
    Chung and Zisserman~\cite{Chung16}           & 38.9\% & - \\ \hline 
    \bf{WAS}  (ours)                             & {\bf 23.8\%}   & {\bf 3.0\%} \\ \hline 

\end{tabular} 
\spacetablecapt
\caption{{\bf Word error rates} on external lip reading datasets.}
\label{table:extres}
\vspace{-5pt}
\end{table}

\subsection{GRID dataset} 
\spacesubsection

\begin{figure}[ht]
\centering 
\includegraphics[width=0.32\linewidth]{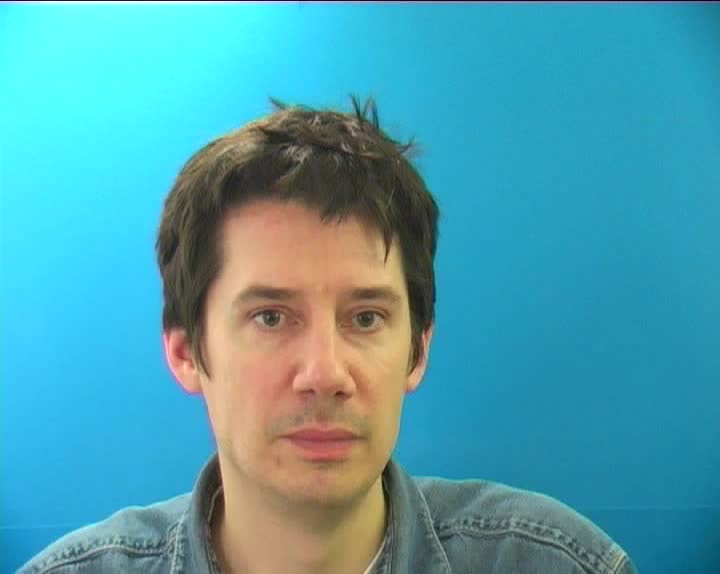}
\includegraphics[width=0.32\linewidth]{images/grid/000002.jpg}
\includegraphics[width=0.32\linewidth]{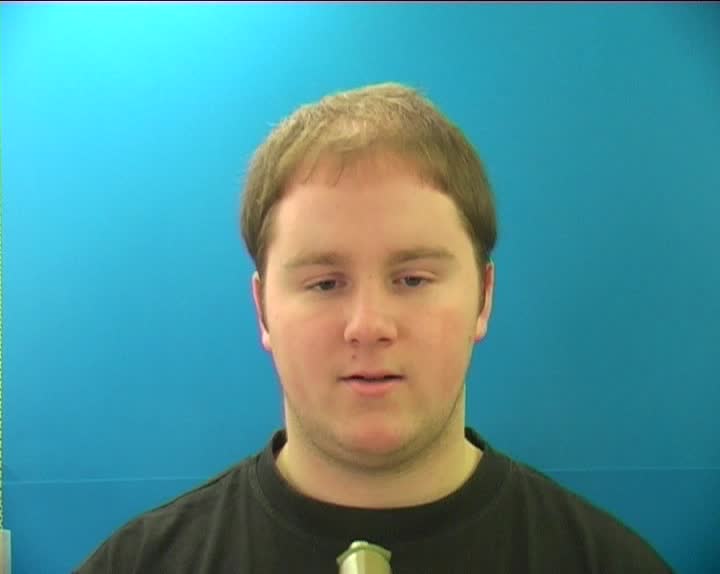}
\caption{Still images from the GRID dataset.
}
\label{fig:grid} 
\vspace{-5pt}
\end{figure}

The GRID dataset~\cite{cooke2006audio} 
consists of 34 subjects,
each uttering 1000 phrases.
The utterances are single-syntax multi-word sequences of 
\texttt{verb} (4) + \texttt{color} (4) + \texttt{preposition} (4)
+ \texttt{alphabet} (25) + \texttt{digit} (10) + \texttt{adverb} (4) ;
{\em e.g.} `put blue at A 1 now'. 
The total vocabulary size is 51, but the number of possibilities
at any given point in the output is effectively constrained to the numbers in the brackets above.
The videos are recorded in a controlled lab environment, shown in Figure~\ref{fig:grid}.

\noindent{\bf Evaluation protocol. } 
The evaluation follows the standard protocol of
\cite{wand2016lipreading} and \cite{yannis2016lipnet} --
the data is randomly divided 
into train, validation and test sets, where 
the latter contains 255 utterances for each speaker.
We report the word error rates.
Some of the previous works report
word accuracies, which is defined as
 $(\mathtt{WAcc} = 1-\mathtt{WER})$.

\noindent{\bf Results.}
The network is fine-tuned for one epoch on the GRID dataset training set.
As can be seen in Table~\ref{table:extres},
our method achieves a
strong performance of 3.0\% (WER), that substantially 
exceeds the current state-of-the-art.



\section{Summary and extensions}
\label{sec:conc}
\spacesection

In this paper, we have introduced the {\it `Watch, Listen, Attend and Spell'}
network model that can transcribe speech into characters.  The model utilises
a novel dual attention mechanism that can operate over visual input
only, audio input only, or both.  Using this architecture, we
demonstrate lip reading performance that beats a professional lip
reader on videos from BBC television.  The model also surpasses the
performance of all previous work on standard lip reading benchmark
datasets, and we also demonstrate that visual
information helps to improve speech recognition performance even when
the audio is used.

There are several interesting extensions to consider: first, the
attention mechanism that provides the alignment is unconstrained, yet
in fact always must move monotonically from left to right. This
monotonicity could be incorporated as a soft or hard constraint;
second, the sequence to sequence model is used in batch mode --
decoding a sentence given the entire corresponding lip
sequence. Instead, a more on-line architecture could be used, where
the decoder does not have access to the part of the lip sequence in
the future; finally, it is possible that research of this type could
discern important discriminative cues that are beneficial for teaching
lip reading to the deaf or hard of hearing.

\ifcvprfinal
\noindent\textbf{Acknowledgements.} Funding for this research is provided by the EPSRC 
Programme Grant Seebibyte EP/M013774/1. We are very grateful to Rob Cooper and Matt 
Haynes at BBC Research for help in obtaining the dataset.
\fi

{\small
\bibliographystyle{ieee}
\bibliography{shortstrings,vgg_local,vgg_other,mybib}
}

\newpage
\quad
\newpage

\section{Appendix}

\subsection{Data visualisation}

$\texttt{https://youtu.be/5aogzAUPilE}$ shows visualisations of
 the data and the model predictions.
All of the captions at the bottom of the video
 are predictions generated by the
`{\it Watch, Attend and Spell}' model.
The video-to-text alignment is shown by
the changing colour of the text,
and is generated
by the attention mechanism.

\subsection{The ConvNet architecture}

\begin{table}[ht]
\scriptsize
\centering
\begin{tabular}{| c | c |  c | c | c | c | }
  \hline
  \textbf{Layer}  &  \textbf{Support} & \textbf{Filt dim.} & \textbf{\# filts.} 
  & \textbf{Stride} & \textbf{Data size} \\ \hline 

  conv1 & 3$\times$3 & 5  & 96 & 1$\times$1 & 109$\times$109 \\ \hline
  pool1 & 3$\times$3 & -  & - & 2$\times$2 & 54$\times$54 \\ \hline 
  conv2 & 3$\times$3 & 96  & 256 & 2$\times$2 & 27$\times$27 \\ \hline
  pool2 & 3$\times$3 & -  & - & 2$\times$2 & 13$\times$13 \\ \hline   
  conv3 & 3$\times$3 & 256  & 512 & 1$\times$1 & 13$\times$13 \\ \hline
  conv4 & 3$\times$3 & 512  & 512 & 1$\times$1 & 13$\times$13 \\ \hline
  conv5 & 3$\times$3 & 512  & 512 & 1$\times$1 & 13$\times$13 \\ \hline
  pool5 & 3$\times$3 & -  & - & 2$\times$2 & 6$\times$6 \\ \hline   
  fc6   & 6$\times$6 & 512  & 512 & - & 1$\times$1 \\ \hline

\end{tabular} 
\normalsize
\vspace{2pt}
\caption{
The ConvNet architecture.
}
\label{table:convnet}
\end{table}

\subsection{The RNN details}

In this paper, we use the standard long short-term memory (LSTM) network
\cite{hochreiter1997long}, which is implemented as follows:

\begin{align}
i_t &= \sigma(W_{xi} x_{t} + W_{hi} h_{t-1} + W_{ci} c_{t-1} + b_{i} ) \\
f_t &= \sigma(W_{xf} x_{t} + W_{hf} h_{t-1} + W_{cf} c_{t-1} + b_{f} ) \\
c_t &= f_{t} c_{t-1} + i_{t} \mathtt{tanh}(W_{xc} x_{t} + W_{hc} h_{t-1} + b_{c}) \\
o_t &= \sigma(W_{xo} x_{t} + W_{ho} h_{t-1} + W_{co} c_{t-1} + b_{o} ) \\
h_t &= o_{t} \mathtt{tanh}(c_{t})
\end{align}

\noindent where $\sigma$ is the sigmoid function, $i, f, o, c$ are the 
input gate, forget gate, output gate and cell activations, all
of which are the same size as the hidden vector $h$.

\begin{figure*}[ht]
\centering 
\includegraphics[width=0.8\textwidth]{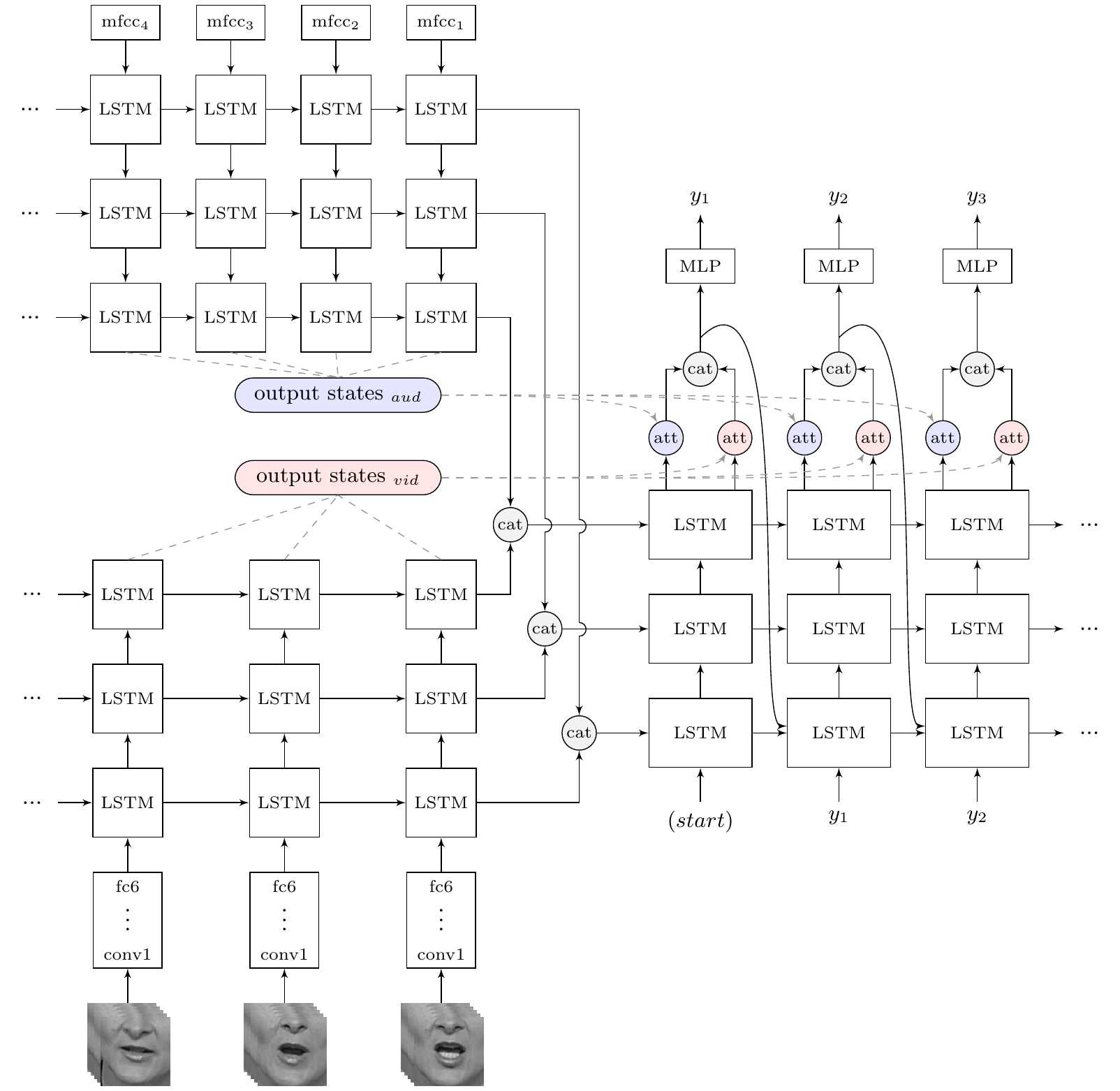}
\caption{Architecture details.}
\label{fig:encdec} 
\end{figure*}

The details of the encoder-decoder architecture is shown
in Figure~\ref{fig:encdec}.

\subsection{The attention mechanism}

This section describes the attention mechanism. 
The implementation is based on the work of
Bahdanau {\em et al.}~\cite{bahdanau2014neural}.
The attention vector $\alpha^v$ for the video stream, 
also often
called the alignment, is computed as follows:

\begin{align}
\alpha^v_{k,i}  &= \mathtt{Attention^v} (s^d_k, \mathbf{o}^v) \\
e_{k,i} &= w^T \mathtt{tanh}(W s^d_k + V o^v_{i} + b) \\
\alpha^v_{k,i}  &= \frac{\mathtt{exp}(e_{k,i})}{\sum\limits_{i=1}^n \mathtt{exp}(e_{k,i})}
\end{align}

where $w$, $b$, $W$ and $V$ are weights to be learnt, and
all other notations are same as in the main paper.

\begin{figure*}[th!]
\centering 
\includegraphics[width=0.45\textwidth]{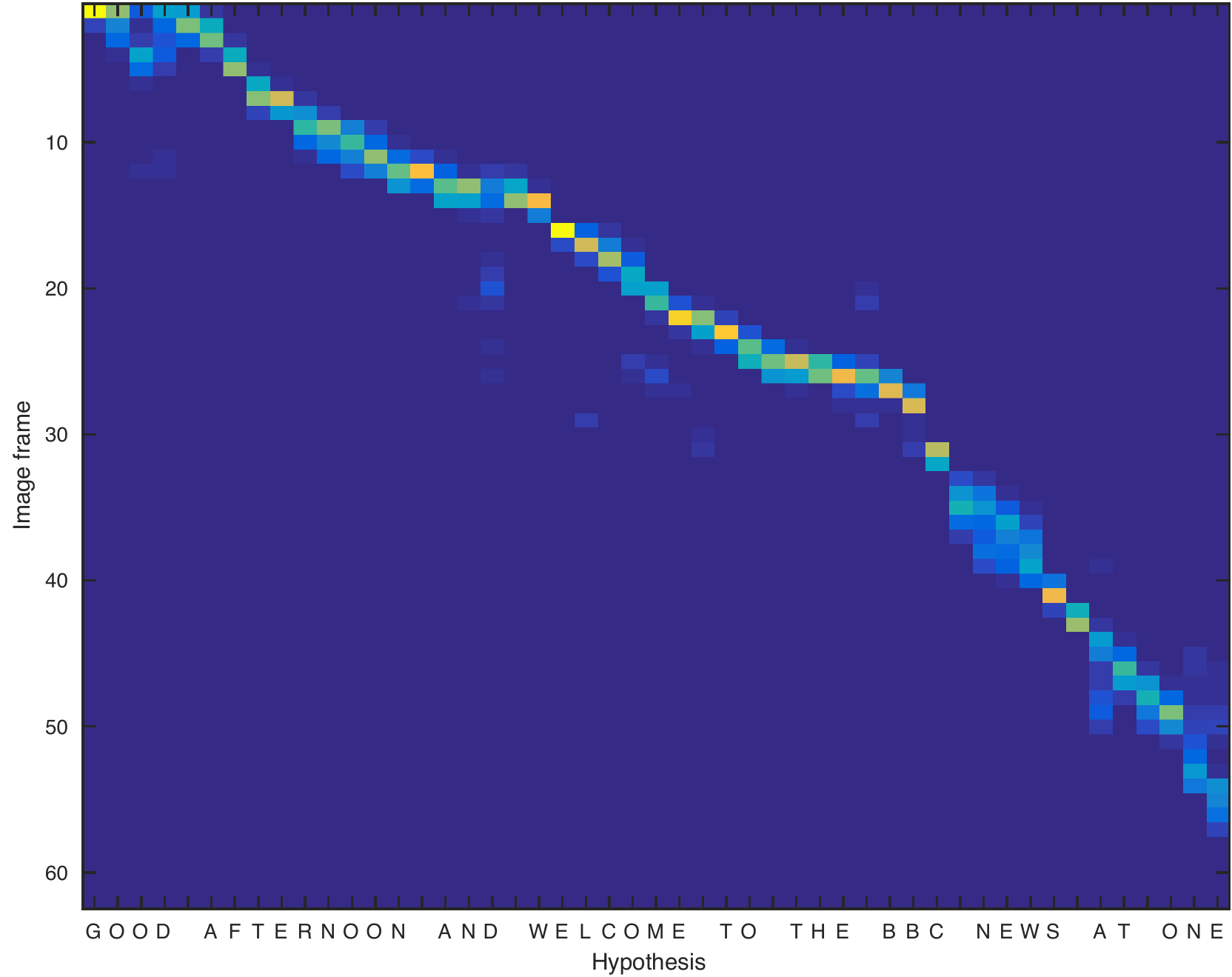}
\includegraphics[width=0.45\textwidth]{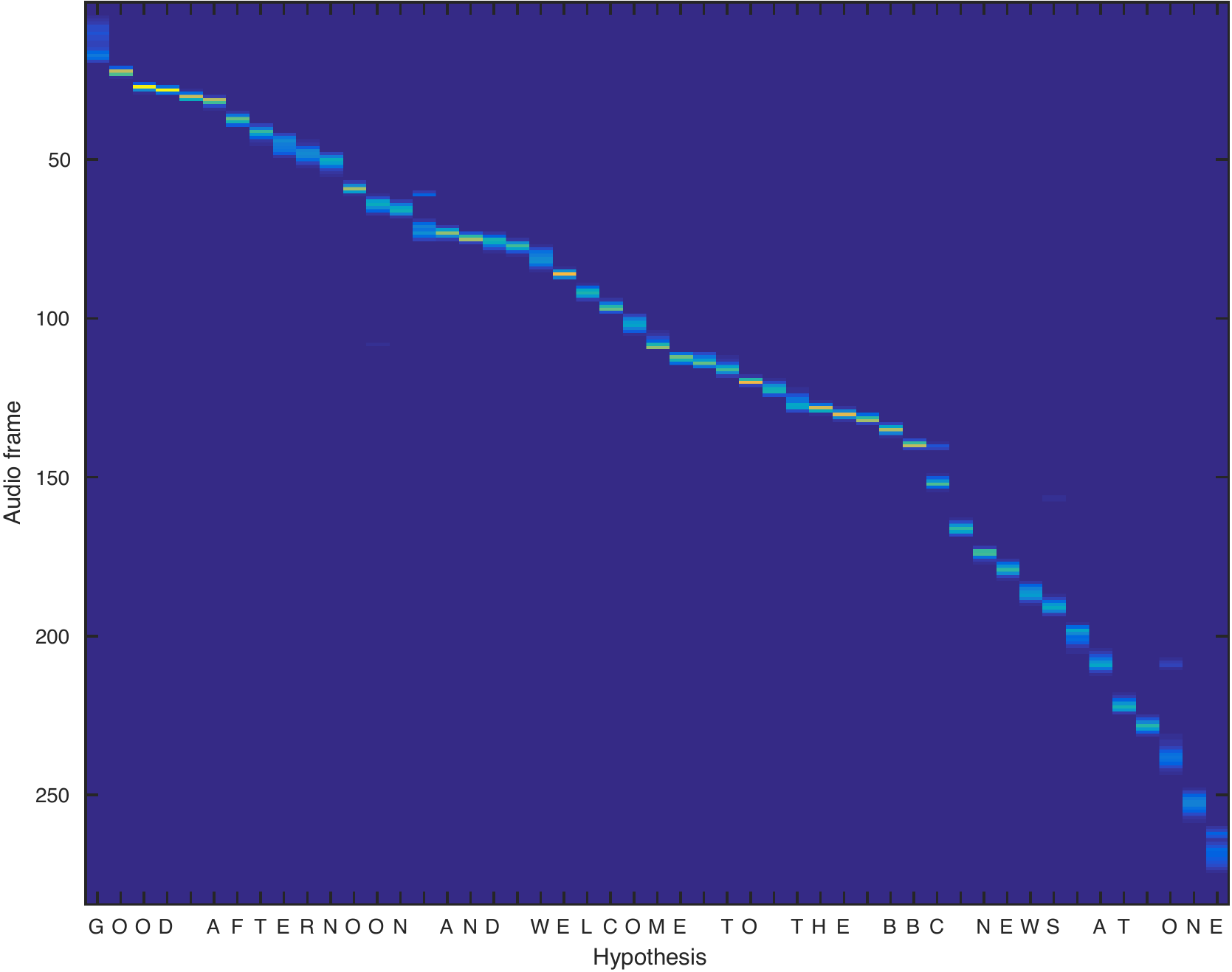}
\caption{Video-to-text (left) and
audio-to-text (right) alignment using
the attention outputs.}
\label{fig:attnvis2} 
\end{figure*}

An example  is shown for both modalities in Figure~\ref{fig:attnvis2}.

\subsection{Model output}

The `Watch, Listen, Attend and Spell' model generates the output sequence
from the list in Table~\ref{table:outputseq}.

\begin{table}[ht]
\centering
\begin{tabular}{| p{7cm} |  }
  \hline
  \texttt {A B C D E F G H I J K L M N O P Q R S T U V W X Y Z 0 1 2 3 4 5 6 7 8 9 , . ! ? : ' [sos] [eos] [pad]} 
  \\ \hline 
\end{tabular} 
\caption{The output characters}
\label{table:outputseq}
\end{table}

\end{document}